\documentclass[runningheads]{llncs}
\usepackage{amsmath}
\usepackage{amssymb}
\usepackage{esvect}
\usepackage[T1]{fontenc}
\usepackage{graphicx}
\usepackage{comment}
\usepackage[table]{xcolor}  
\usepackage{url}
\begin{document}

\title{EgoInertia-MI: A Multimodal Egocentric Vision and IMU Benchmark for Motor Impairment Assessment}
\titlerunning{EgoInertia-MI: Egocentric Vision and IMU for Motor Impairment Analysis}

\author{
Fatemah Alhamdoosh\inst{1}
Pietro Pala\inst{1}
Abduallah Mohamed\inst{2,*} \and
DK Arvind\inst{3}
}

\authorrunning{F. Alhamdoosh et al.}

\institute{
University of Florence, Italy\\
\email{\{fatemah.alhamdoosh,pietro.pala\}@unifi.it}
\and
Meta Reality Labs, USA\\
\email{abduallah.adel.omar@gmail.com}
\and
University of Edinburgh, UK\\
\email{D.Arvind@ed.ac.uk}
}

\maketitle
\begingroup
\renewcommand\thefootnote{*}
\footnotetext{This work was done outside of Meta in a personal capacity.}
\endgroup
\begin{abstract}
Motor impairments, including tremor, bradykinesia, gait abnormalities, and postural instability, are common across many neurological and movement-related conditions. Conventional clinical assessments are often intermittent and may fail to capture subtle temporal variations in motor behavior. While wearable IMUs and third-person video have shown promise for objective motor assessment, third-person recordings raise privacy concerns and require constrained acquisition setups. In contrast, egocentric vision provides a more naturalistic and privacy-aware alternative.
In this work, we introduce EgoInertia-MI, a multimodal benchmark dataset combining synchronized egocentric video and wearable IMU signals for motor impairment analysis. The dataset contains 19 upper- and lower-body activities performed by healthy volunteers simulating varying levels of motor impairment severity levels: no impairment, mild impairment, and severe impairment. We establish two benchmark tasks: action recognition and motor impairment severity estimation, and evaluate multiple unimodal and multimodal baselines. Experimental results show that egocentric video provides strong cues for motor impairment assessment, while multimodal fusion achieves the best overall performance, reaching 0.78 Macro-F1 for severity estimation and 0.93 Macro-F1 for action recognition. These findings highlight the potential of combining egocentric vision and wearable sensing for ecologically valid and privacy-aware motor assessment. Code and data are available at:\url{https://fatemah-alh.github.io/EgoInertia-MI-Page/}.

\end{abstract}

\section{Introduction}
Subtle alterations in movement dynamics, including impaired gait, reduced movement speed, tremor, and instability, are key indicators of functional decline across numerous neurological and movement-related conditions~\cite{Inciarte2022}. Accurate assessment of these impairments is critical for monitoring functional decline, evaluating interventions, and supporting long-term healthcare management. However, conventional clinical evaluations are typically intermittent and may fail to capture subtle temporal fluctuations and variations in motor behavior observed during daily activities. Recent advances in wearable sensing and computer vision have enabled new opportunities for scalable and quantitative analysis of human movement outside controlled clinical environments. Among these technologies, inertial measurement units (IMUs) and vision-based approaches have shown promising capabilities for movement analysis and impairment estimation~\cite{vijayan2021review,sharma2022advancements}. Despite promising progress in vision-based movement analysis, most existing systems rely on third-person recordings acquired using smartphones or controlled multi-camera setups~\cite{vun2024vision,kaur2022vision}. Such configurations often require carefully constrained acquisition environments and frequently capture identifiable information, including facial appearance and full-body imagery, raising important privacy and ethical concerns~\cite{xiang2022being}. Moreover, third-person viewpoints may fail to capture subtle self-motion dynamics and interaction patterns naturally observed during daily activities. These limitations reduce scalability and hinder deployment in ecologically valid real-world settings~\cite{liu2022explicit}.

In contrast, egocentric vision offers a promising alternative for movement assessment by capturing self-motion dynamics, hand-object interactions, and locomotion patterns directly from the first-person perspective. Egocentric motion patterns inherently encode hand dynamics, body stabilization, locomotion behavior, and interaction smoothness, all of which are closely related to motor function. Compared with third-person recordings, egocentric sensing naturally reduces the visibility of identity-sensitive information while enabling more flexible and unobtrusive acquisition during daily activities~\cite{grauman2022ego4d,yang2025egolife}. Despite the growing success of egocentric vision in activity understanding, its potential for fine-grained motor impairment assessment remains largely unexplored. Wearable IMU sensing offers temporally precise and quantitatively grounded measurements of motion dynamics, capturing subtle variations in acceleration, rhythm, and movement stability that are often difficult to observe visually alone \cite{dirgova2022wearable,ordonez2016deep}. In contrast, egocentric vision situates these inertial patterns within their behavioral and environmental context, disambiguating motion through visual evidence of user activity and interaction dynamics. 
 The development of public datasets in this domain are constrained by substantial ethical, logistical, and privacy challenges associated with acquiring data from clinical populations. This result in limited-scale datasets with restricted accessibility, reducing their utility for developing and benchmarking modern machine learning approaches.
To address these challenges, we introduce EgoInertia-MI, a multi-modal benchmark designed to investigate the synergy between IMU and egocentric vision in motor impairment assessment. The dataset consists of recordings from healthy volunteers instructed to mimic a range of motor impairment characteristics at different severity levels. It contains 19 upper- and lower-body activities performed across three conditions: no impairment, mild impairment, and severe impairment. Data are acquired using chest-mounted egocentric cameras together with wearable IMU sensors.

\begin{figure}
\centering
\includegraphics[
        width=0.90\linewidth,
        trim={0cm 0cm 0cm 0cm},
        clip
    ]{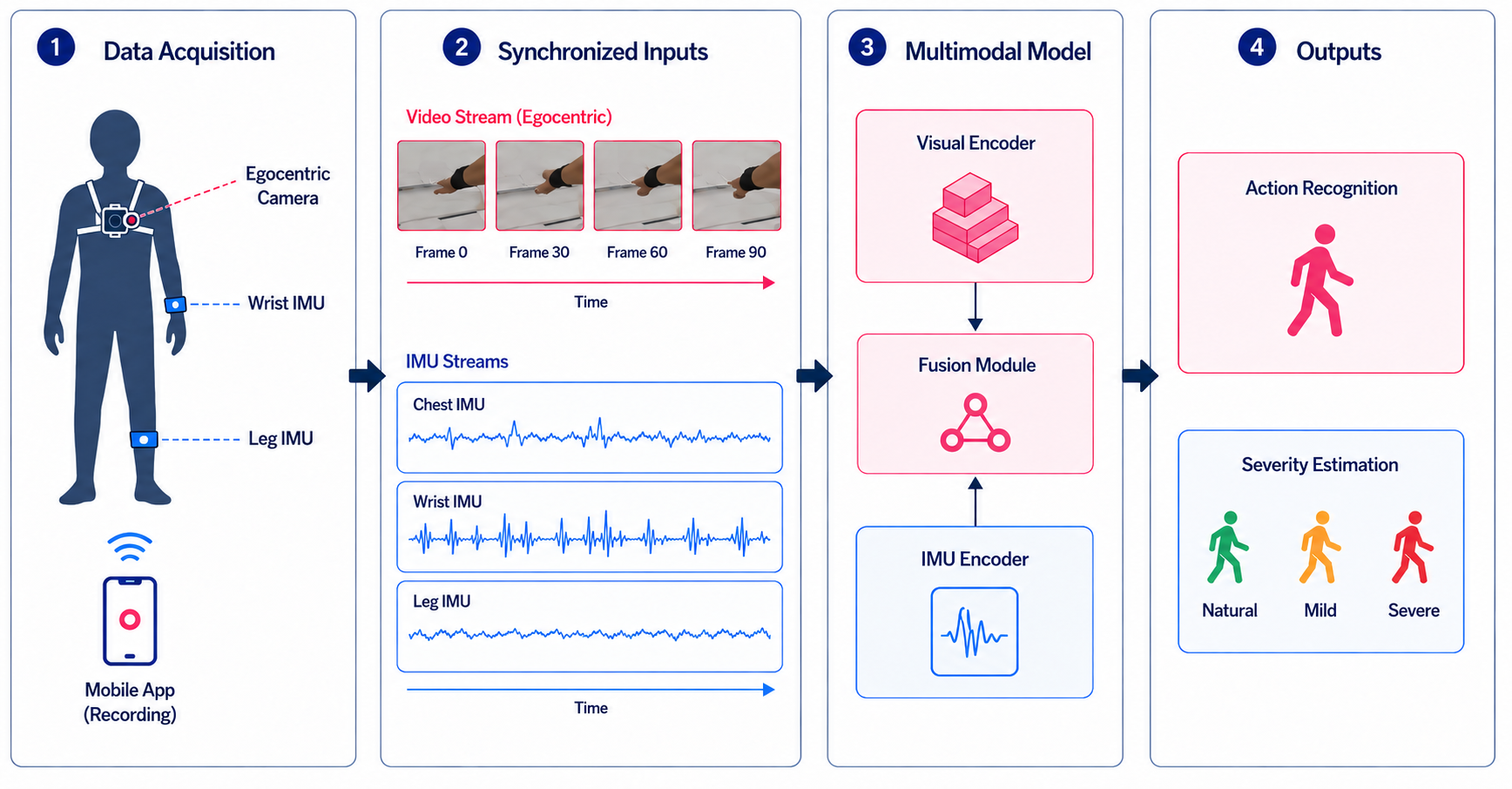}
\caption{Overview of EgoInertia-MI. Synchronized egocentric video and wearable IMU streams are processed by a multimodal model for movement understanding and impairment severity estimation.}
\label{fig:overall}
\end{figure}

In addition to the dataset, we establish two benchmark tasks: action recognition and motor impairment severity estimation. We further provide standardized evaluation protocols and baseline experiments across unimodal and multimodal architectures. Figure~\ref{fig:overall} presents an overview of the proposed EgoInertia-MI framework, including multimodal acquisition, synchronized egocentric and inertial streams, and downstream movement analysis tasks. Experimental results demonstrate that egocentric vision encodes strong predictive cues for movement impairment analysis, while multimodal fusion consistently achieves the best overall performance across both benchmark tasks.
The main contributions of this work are summarized as follows:
\begin{itemize}
\item We introduce \textit{EgoInertia-MI}, a synchronized multimodal benchmark combining egocentric vision and wearable IMU sensing for fine-grained motor impairment assessment across 19 activities and three severity levels.

\item We define two benchmark tasks, action recognition and impairment severity estimation, with standardized evaluation protocols across unimodal and multimodal learning settings.

\item We demonstrate that egocentric vision provides strong cues for motor impairment analysis, while multimodal fusion consistently achieves the best overall performance, highlighting the complementary nature of visual and inertial sensing.
\end{itemize}

\section{Related Work}

\subsection{Wearable Sensing for Motor Impairment Analysis}

Wearable inertial sensors have become a widely adopted solution for quantitative analysis of human motion due to their low cost, portability, and high temporal resolution. IMU sensing is extensively used in general human activity recognition (HAR), where large-scale benchmarks and deep learning methods have significantly advanced wearable-based action understanding~\cite{bock2024wear}. Recent progress further includes foundation and self-supervised models for wearable sensing, enabling transferable representations across diverse activities and sensor configurations~\cite{narayanswamy2025scaling}. Within movement and neurological assessment, prior work has explored IMU-based analysis of gait abnormalities, tremor, bradykinesia, and postural instability~\cite{delrobaei2017using,vijayan2021review}. Early approaches mainly relied on handcrafted temporal and frequency-domain features, while recent studies increasingly employ convolutional, recurrent, and transformer-based architectures to directly model raw inertial sequences~\cite{moon2017monitoring,varghese2024pads,anderson2026weargait,voisard2025dataset}. 
\subsection{Vision-Based Motor Assessment}

Vision-based approaches have recently emerged as a promising direction for automated motor assessment. Advances in pose estimation and deep learning have enabled quantitative analysis of gait patterns, hand movements, tremor, and clinical motor scores from video recordings~\cite{lu2020vision}. Several studies have demonstrated the feasibility of estimating movement severity and clinical assessment scales directly from RGB videos or extracted skeletal representations~\cite{robinson2025strokevision}. 
However, most existing methods rely on third-person recordings acquired in controlled environments using smartphones or multi-camera systems~\cite{ranjan2025computer,acevedo2023video}. Such approaches frequently capture identifiable visual information, including facial appearance and full-body imagery, raising privacy and ethical concerns. In addition, third-person viewpoints may overlook subtle self-motion dynamics and interaction patterns that are naturally observable from the first-person perspective.
\subsection{Multimodal and Egocentric Human Activity Understanding}

Egocentric vision has recently gained significant attention in activity understanding due to its ability to capture first-person interactions, self-motion patterns, and user-centric contextual information~\cite{grauman2022ego4d,yang2025egolife,feichtenhofer2020x3d}. Large-scale datasets and benchmarks have accelerated research in egocentric perception for daily activity recognition, interaction modeling, and multimodal learning. In parallel, recent studies have explored the complementarity between wearable inertial sensing and egocentric video for human activity recognition tasks~\cite{gong2023mmg}. 
Nevertheless, the use of egocentric vision for motor impairment assessment remains largely unexplored. Existing datasets and benchmarks in movement analysis primarily rely on wearable sensing or third-person recordings, with no publicly available benchmark specifically designed to evaluate synchronized egocentric vision and wearable IMU sensing for fine-grained motor impairment analysis~\cite{pintea2018hand,morgan2023multimodal}. Our work addresses this gap through the introduction of EgoInertia-MI, a benchmark dataset and evaluation framework for multimodal motor impairment assessment.

\section{Dataset and Methodology}

\subsection{Study Design and Ethics}

The dataset was designed to capture multimodal recordings of motor activities under controlled levels of simulated impairment. Ethical approval for this study was granted by the ethics committee of the School of Informatics, University of Edinburgh (Application No. 147186), and all procedures adhered to established guidelines for human subject research. A total of 17 healthy volunteers participated in the study. Participants were selected to perform a structured set of activities designed to reflect declined motor functions, while maintaining a balance between controlled experimental conditions and naturalistic behavior. All participants provided informed consent. Each participant was assigned a unique de-identified ID, and no personally identifiable information was stored.

\subsection{Data Acquisition Setup}

The data collection setup consisted of two wearable Respeck IMU sensors~\cite{arvind2019characterisation} and a chest-mounted camera. The IMU sensors were positioned on the wrist and lower leg (shin) to capture upper- and lower-limb motion, recording accelerometer and gyroscope data at 25.5 Hz. This sampling rate corresponds to a Nyquist frequency of 12.75 Hz, covering common pathological tremor ranges such as Parkinsonian tremor ($\sim$4--7 Hz) and essential tremor ($\sim$4--12 Hz), while supporting energy-efficient wearable data collection. Sensor placement was adjusted according to the participant’s dominant side to ensure natural movement execution. 
Egocentric video was recorded using a GoPro HERO12 mounted on a chest harness, capturing the participant’s first-person perspective at a resolution of ($2160 \times 3840$) and 30 FPS. In addition to video, the camera provides embedded inertial measurements (accelerometer and gyroscope), which were extracted to capture chest and upper-body motion, complementing the wearable IMU signals.

\subsection{Experimental Protocol}

Data collection followed a structured protocol to ensure consistency across participants. Prior to recording, participants received an information sheet and detailed instructions describing the study objectives and procedures. Upon arrival, each participant attended a guided tutorial session lasting approximately 20--30 minutes, during which the researcher explained the protocol and demonstrated example executions of the activities, including different impairment levels. Participants were then assisted in wearing and adjusting the sensing devices, followed by a short calibration recording to verify correct sensor placement and data quality. The recording sessions were conducted in indoor office environments, with nearby fire stairs used for gait-related activities such as walking and stair navigation. To reduce fatigue, the protocol was divided into approximately three consecutive sessions. Participants were given a minimum rest period of 30 seconds between activities, with additional rest provided upon request.

\subsection{Activity Design and Severity Modeling}
\begin{figure}
\centering
\includegraphics[
        width=0.90\linewidth,
        trim={2.5cm 0cm 2.5cm 0cm},
        clip
    ]{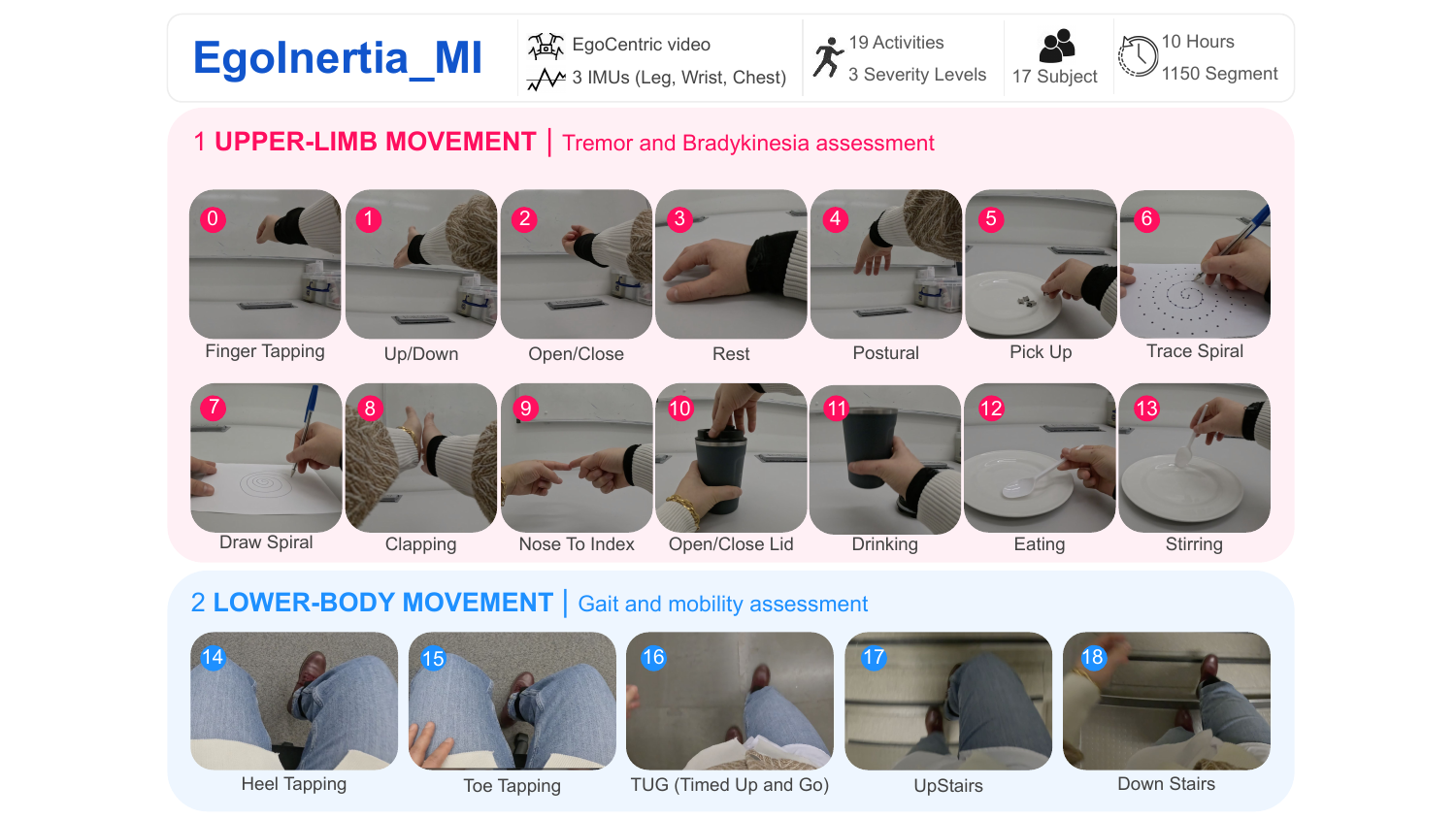}
\caption{Overview of the EgoInertia-MI dataset, showing 19 upper- and lower-body activities with representative egocentric frames and activity labels.}
\label{fig:activities}
\end{figure}

\begin{figure}
\centering
\includegraphics[
        width=\linewidth,
        trim={0cm 4.5cm 0cm 5cm},
        clip
    ]{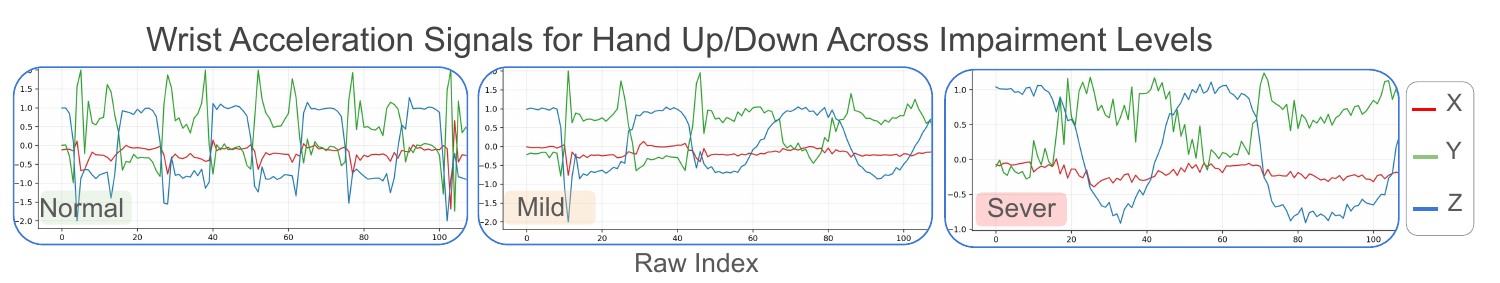}
\caption{Representative wrist IMU acceleration signals for the Hand Up/Down task across different impairment levels.}
\label{fig:imu_sig}
\end{figure}

Figure~\ref{fig:activities} provides an overview of the activity set, which is organized into two main categories: upper-limb activities aimed at characterizing tremor and bradykinesia-related movement patterns, and lower-body activities focused on gait and mobility assessment. The protocol combines clinically established motor evaluations, such as finger tapping and the Timed Up and Go (TUG) test, with representative activities of daily living, thereby balancing controlled clinical assessment with ecological validity.
Each activity was performed under three levels of simulated motor severity to capture progressive variations in movement quality and impairment intensity. The first condition corresponded to \textbf{natural} execution, reflecting unconstrained and typical movement patterns. The second represented \textbf{mild impairment}, characterized by slight reductions in movement speed and coordination, accompanied by subtle tremor or instability. The third simulated \textbf{severe impairment}, involving pronounced motor degradation, including markedly reduced movement speed, increased tremor amplitude, impaired balance, and greater difficulty in task execution. Figure \ref{fig:imu_sig} illustrates a representative Hand Up/Down sequence, showing how increasing impairment severity is reflected by progressively slower and less stable motion patterns. This protocol design balances clinical relevance with the need for controlled simulation and consistent annotation of motor impairment across activities. Furthermore, requiring participants to perform the full activity set under multiple severity conditions enables the dataset to capture intra-subject variability that resembles fluctuations commonly observed in clinical practice, such as changes associated with treatment response or medication ON/OFF states

\subsection{Annotation and Synchronization}
Each recording session lasts approximately 20 minutes and comprises multiple activities performed sequentially under the predefined protocol conditions. For every session, three synchronized data streams are acquired: (i) egocentric video with embedded camera IMU measurements, (ii) wrist-mounted IMU signals, and (iii) leg-mounted IMU signals. Synchronization between modalities is achieved using the recording start timestamps of each device, which are used as reference anchors for temporal alignment.
Video recordings are segmented and annotated using \textit{LosslessCut}. For each segment, the corresponding start and end timestamps are recorded, together with a set of structured annotations including the activity label (0--18), severity level (0--2), active hand configuration (L: left, R: right, RL: both hands), and sensor placement configuration (RR: right wrist and right leg, LL: left wrist and left leg). These annotations, combined with the synchronization reference timestamps, are subsequently used to temporally segment and align the IMU streams through relative timing offsets. Each final multimodal sample therefore consists of a synchronized video clip paired with an IMU feature matrix of size $T \times 18$, where $T$ denotes the number of temporal samples within the segment.

\subsection{Dataset Overview}

\begin{figure}
\includegraphics[
        width=\linewidth,
        trim={0cm 6cm 0cm 7cm},
        clip
    ]{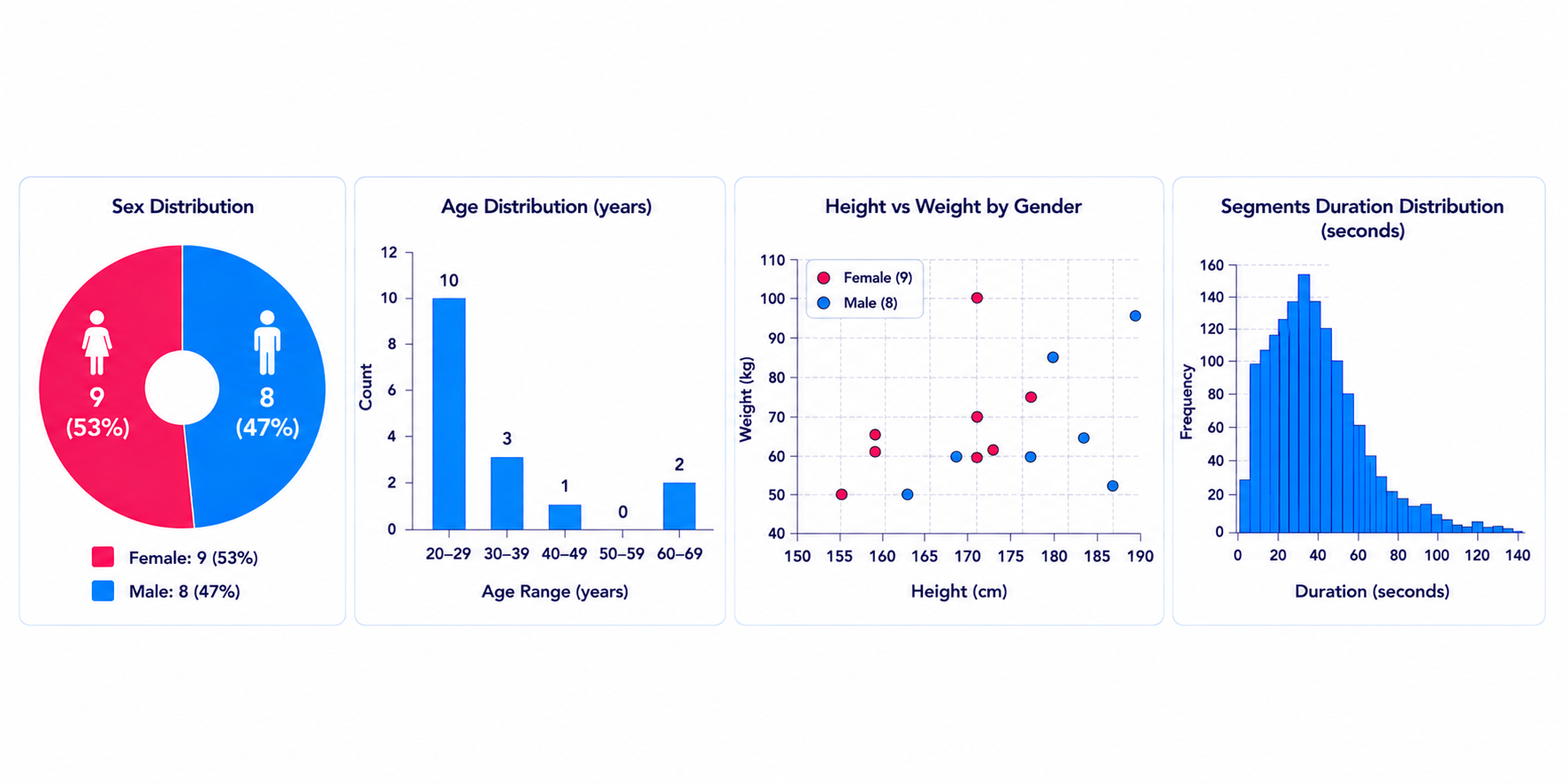}
\caption{Demographic and recording statistics of EgoInertia-MI dataset, including sex and age distributions, anthropometric measurements, and segment duration characteristics.}
\label{fig:dataset_info}
\end{figure}

The proposed dataset comprises approximately 1,150 annotated multimodal segments collected from 17 participants (9 females and 8 males), corresponding to nearly 10 hours of synchronized recordings. Participants span a broad demographic range, with ages between 20 and 69 years (mean: 33.2 years, standard deviation: 14.6 years), contributing to increased inter-subject variability and improved ecological validity. The dataset includes 19 activities acquired using two body-worn IMUs (wrist and leg) and a wearable camera. Annotations are provided across three severity levels (0--2), with Severity 0 accounting for approximately 2 h 26 min of recordings, Severity 1 for 3 h 20 min, and Severity 2 for 4 h 13 min. Segment durations vary substantially depending on the performed task. Repetitive motor activities typically last between 25 and 40 seconds, whereas unconstrained tasks, such as walking or drawing, are recorded until task completion. Consequently, segment durations range from approximately 4 to 140 seconds, with an average duration of 33.6 seconds, reflecting the heterogeneous and realistic nature of clinical motor assessments.
Figure~\ref{fig:dataset_info} illustrates the demographic distributions and recording statistics.

\section{Benchmark Tasks and Evaluation Protocol}

The proposed EgoInertia-MI dataset is designed to support multimodal learning from wearable inertial signals and egocentric video for movement analysis. The benchmark includes both unimodal and multimodal settings, enabling the evaluation of models operating on IMU data, video data, or their combination. This design reflects realistic deployment scenarios in which one or more modalities may be unavailable or corrupted.

We define two primary benchmark tasks:

\paragraph{Impairment Severity Estimation.}
The primary task consists of estimating movement impairment severity from wearable IMU signals, egocentric video, or both modalities jointly. Severity estimation is formulated as a multi-class classification problem with three severity levels (0--2). Since movement impairments are inherently context-dependent, similar severity levels may manifest differently across activities and motion types. Consequently, this task requires models not only to capture low-level motion characteristics, but also to develop contextual understanding of activity-dependent movement patterns. The benchmark therefore encourages the development of robust multimodal systems capable of modeling subtle variations in movement quality across diverse daily living and clinically relevant actions. 
\paragraph{Action Recognition.}
In addition to severity estimation, the benchmark includes fine-grained action recognition across 19 daily living and clinically relevant activities. The objective of this task is to recognize actions under varying movement impairment conditions, requiring models to remain robust to substantial intra-class variations in motion execution. Unlike conventional action recognition benchmarks that predominantly capture normative movement patterns, EgoInertia-MI introduces significant variability in motion dynamics, execution speed, coordination, and movement quality across subjects and severity levels. This setting promotes the development of more inclusive and generalizable action recognition systems capable of learning activity-specific representations that remain reliable under diverse movement characteristics and motor impairments.
\paragraph{Evaluation Protocol.}
To ensure robust subject-independent evaluation, all experiments follow a 5-fold cross-validation protocol with subject-disjoint splits, preventing any identity leakage between training, validation, and test sets. For each fold, data from 3 subjects are held out for testing, while the remaining subjects are divided into training and validation subsets comprising 12 and 2 subjects, respectively. Performance is primarily evaluated using Macro F1-score, which provides a balanced assessment under multi-class and class-imbalanced settings.
\section{Benchmark Baselines}
We evaluate representative architectures spanning temporal sequence modeling, video understanding, and multimodal fusion to establish benchmark performance on EgoInertia-MI.
Given a multimodal sample $S_n=\{v_n, i_n\}$, where $v_n$ and $i_n$ denote the synchronized video and IMU streams respectively, each recording is segmented into non-overlapping clips of fixed duration. Since sequence lengths vary between approximately 4 and 140 seconds, all recordings $S_n$ are divided into 5-second multimodal clips $S_n \rightarrow \{C_{n1}, C_{n2}, \dots, C_{nt}\}$, where each clip $C_{nt}=\{v_{nt}, i_{nt}\}$ inherits the labels associated with the original recording. Clips shorter than 5 seconds are padded using the last available temporal observations. During inference, recording-level predictions are obtained by averaging clip-level logits across all clips associated with the same recording.

\subsection{IMU-Based Architectures}
For IMU-based experiments, each input clip is represented as a temporal feature matrix $i_{nt} \in \mathbb{R}^{T \times D}$, where $T=127$ corresponds to the temporal dimension of a 5-second window and $D=18$ denotes the inertial feature dimension. The representation concatenates tri-axial accelerometer and gyroscope measurements acquired from wrist, leg, and chest sensors. To model temporal dynamics, we evaluate several sequence learning architectures commonly used in wearable sensing and human activity recognition, including CNN, LSTM, TCN~\cite{bai2018empirical}, DeepConvLSTM~\cite{ordonez2016deep}, and HARTransformer~\cite{dirgova2022wearable}. 
\subsection{Video-Based Architectures}

For video-based experiments, each clip contains approximately 150 RGB frames extracted from egocentric recordings. Following standard practice in video understanding, frames are temporally subsampled according to the input requirements of each architecture. Each video clip is therefore represented as a tensor: $v_{nt} \in \mathbb{R}^{ 3 \times  B \times F \times F}$,
where $B \in \{13,16,32\}$ denotes the number of sampled frames and $F \in \{182,256\}$ represents the spatial input resolution used by the corresponding architecture. Since the original recordings are acquired at high spatial resolution ($2160 \times 3840$), frames are resized by scaling the shortest side to either 182 or 256 pixels, followed by center cropping to obtain square spatial inputs while reducing computational complexity. We evaluate representative architectures such as X3D~\cite{feichtenhofer2020x3d}, SlowFast~\cite{feichtenhofer2019slowfast} and V-JEPA ~\cite{assran2025v}, under frozen and fine-tuned training protocols. Detailed input configurations and results are summarized in Table~\ref{tab:benchmark_results}.
\subsection{Multimodal Fusion}

For multimodal learning, we leverage the pretrained unimodal encoders obtained from the IMU-based and video-based experiments. Specifically, temporal clip-level embeddings extracted from the video and IMU branches are denoted as $h(v_{nt})\in \mathbb{R}^{T_v \times 2048}$ and $f(i_{nt}) \in \mathbb{R}^{T_i \times 64}$, respectively, with $T_v$ and $T_i$ representing the temporal lengths of the video and IMU embeddings.

We evaluate late fusion and temporal cross-attention fusion. In late fusion, temporal embeddings from each modality are aggregated using mean pooling, projected into a shared latent space, concatenated, and passed to the classification head. In temporal cross-attention fusion, the projected temporal embeddings are fused before pooling by allowing each modality to attend to the other over time, enabling video features to incorporate inertial dynamics and IMU features to incorporate visual motion cues before final aggregation and prediction.

\section{Results Analysis and Discussion}
\begin{table*}[t]
\centering
\caption{Comparison of IMU-based, video-based, and multimodal models for severity estimation and action recognition. Results are reported as Macro-F1 scores ($\mu \pm \sigma$) across cross-validation folds.}
\label{tab:benchmark_results}
\setlength{\tabcolsep}{4pt}
\renewcommand{\arraystretch}{1.12}
\resizebox{\textwidth}{!}{
\begin{tabular}{llcccc}
\hline
\textbf{Modality} & \textbf{Model} & \textbf{Fine-tuning} & \textbf{Input Dim.} & \textbf{Severity $\uparrow$} & \textbf{Action $\uparrow$} \\
\hline

\textbf{IMU} & TCN~\cite{bai2018empirical} & -- & 127$\times$18 & 0.623 $\pm$ 0.048 & 0.433 $\pm$ 0.069 \\
& LSTM & -- & 127$\times$18 & 0.603 $\pm$ 0.080 & 0.464 $\pm$ 0.059 \\
& CNN & -- & 127$\times$18 & 0.658 $\pm$ 0.059 & 0.561 $\pm$ 0.063 \\
& HARTransformer~\cite{dirgova2022wearable} & -- & 127$\times$18 & 0.625 $\pm$ 0.064 & \textbf{0.568 $\pm$ 0.076} \\
& DeepConvLSTM~\cite{ordonez2016deep} & -- & 127$\times$18 & \textbf{0.689 $\pm$ 0.056} & 0.507 $\pm$ 0.039 \\

\hline

\textbf{Video} & x3d\_xs~\cite{feichtenhofer2020x3d} & False & 13$\times$182$\times$182 & 0.506 $\pm$ 0.057 & 0.807 $\pm$ 0.054 \\
& x3d\_xs & True & 13$\times$182$\times$182 & 0.661 $\pm$ 0.043 & 0.880 $\pm$ 0.022 \\
& x3d\_s & False & 13$\times$182$\times$182 & 0.589 $\pm$ 0.031 & 0.836 $\pm$ 0.026 \\
& x3d\_s & True & 13$\times$182$\times$182 & 0.674 $\pm$ 0.057 & 0.890 $\pm$ 0.014 \\
& x3d\_m & False & 16$\times$256$\times$256 & 0.610 $\pm$ 0.056 & 0.842 $\pm$ 0.047 \\
& x3d\_m & True & 16$\times$256$\times$256 & 0.693 $\pm$ 0.038 & 0.815 $\pm$ 0.061 \\

& vjepa-l~\cite{assran2025v} &False & 32$\times$256$\times$256 & 
0.668 $\pm$ 0.038 & 0.841 $\pm$ 0.042 \\

& slowfast\_r50~\cite{feichtenhofer2019slowfast} & False & 32$\times$256$\times$256 & 0.694 $\pm$ 0.047 & 0.836 $\pm$ 0.012 \\
& slowfast\_r50 & True & 32$\times$256$\times$256 & \textbf{0.750 $\pm$ 0.028} & \textbf{0.925 $\pm$ 0.013} \\

\hline

 \textbf{Multimodal} & CrossAttention & -- & IMU + Video & 0.780 $\pm$ 0.049 & 0.920 $\pm$0.02  \\
 
 & LateFusion & -- & IMU + Video & \textbf{0.784 $\pm$ 0.049} & \textbf{0.934 $\pm$ 0.013} \\

\hline
\end{tabular}}
\end{table*}

\begin{figure}
\centering
\includegraphics[
        width=\linewidth,
        trim={1cm 6cm 1cm 6.5cm},
        clip
    ]{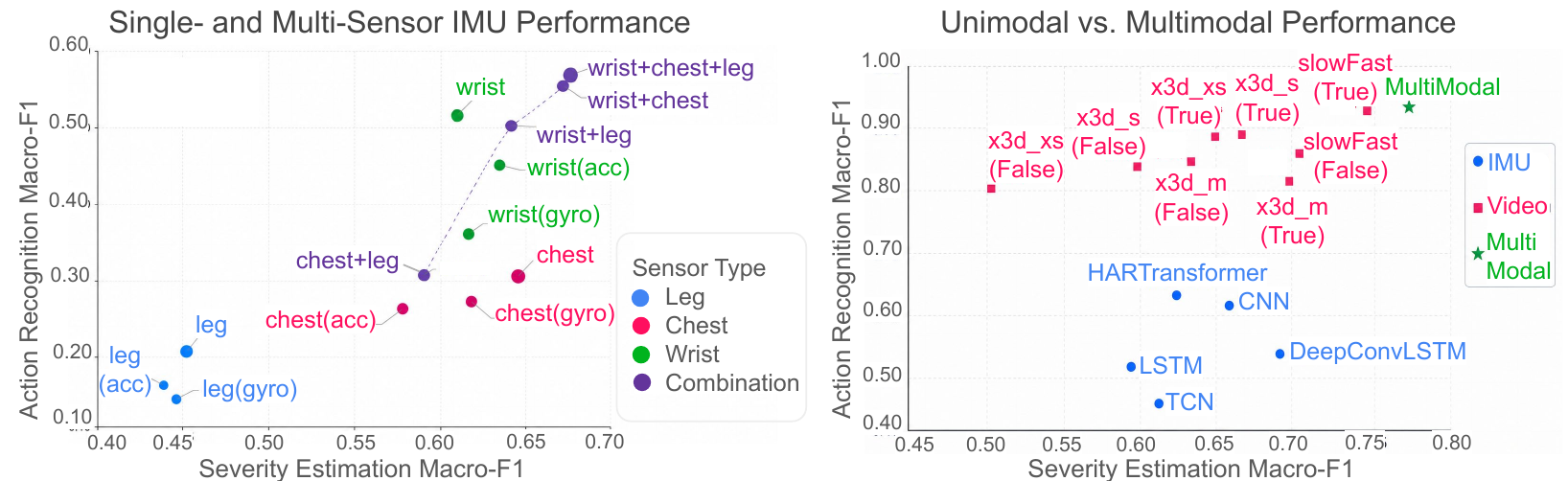}
\caption{Comparative analysis of sensor configurations and learning architectures for severity estimation and action recognition. Left: performance of different IMU sensor combinations. Right: comparison of IMU-based, video-based, and multimodal architectures.}
\label{fig:model_summary}
\end{figure}

\subsection{Unimodal and Multimodal Performance Comparison}

Table~\ref{tab:benchmark_results} presents the quantitative comparison of IMU-based, video-based, and multimodal architectures for severity estimation and action recognition. 

\begin{figure}
\includegraphics[width=\textwidth]{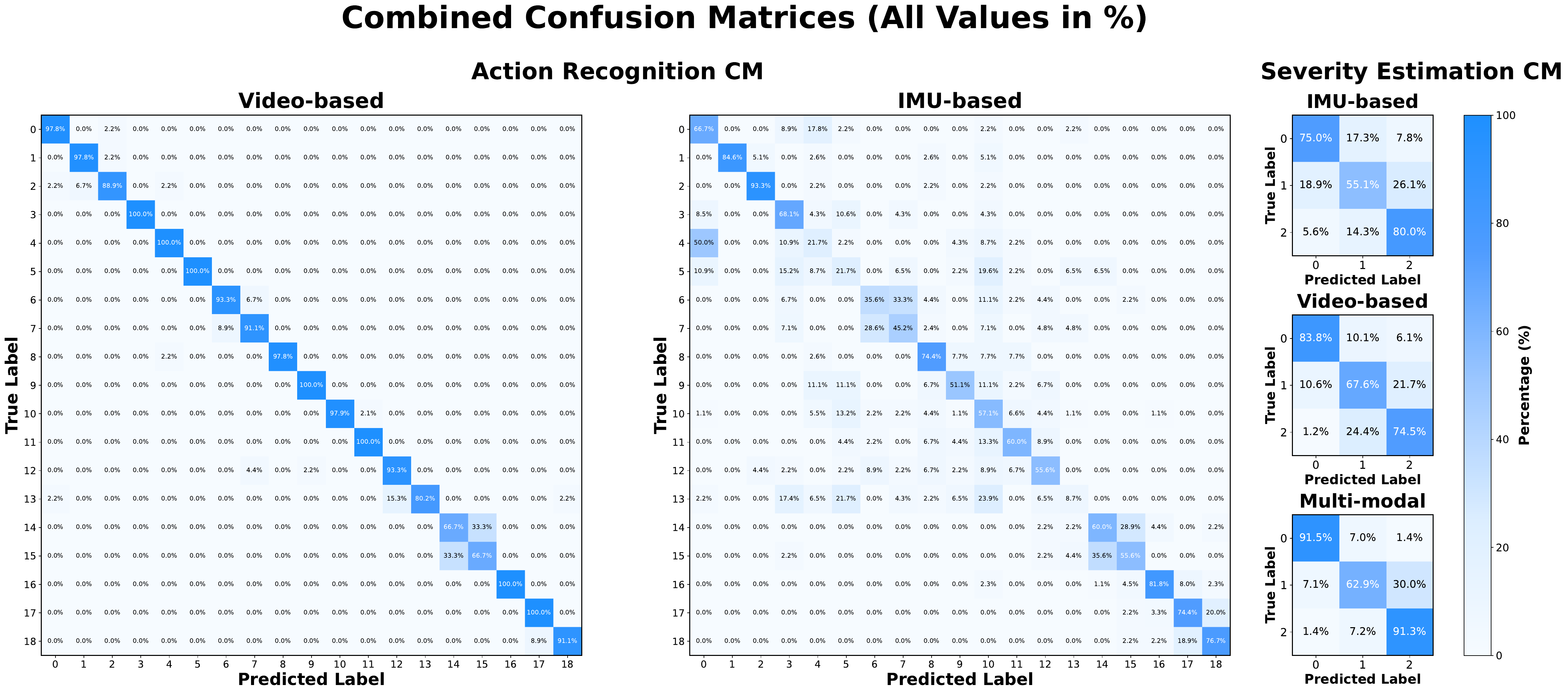}
\caption{Comparison of normalized confusion matrices for action recognition and severity estimation using IMU-based and video-based architectures. Values are reported as row-wise percentages.}
\label{fig:cm}
\end{figure}

Figure~\ref{fig:model_summary}~(Right) further visualizes the relationship between both tasks across the evaluated approaches. IMU-based models achieved competitive severity estimation performance, reaching up to $0.689 \pm 0.056$ Macro-F1, indicating that inertial signals effectively capture fine-grained temporal motion dynamics and biomechanical patterns. However, IMU-based methods achieved lower action recognition performance ($0.568 \pm 0.076$ Macro-F1), suggesting that wearable signals alone may lack sufficient contextual and spatial information to reliably distinguish wide range of activities. Video-based approaches substantially improved performance in both tasks, with the fine-tuned SlowFast architecture achieving $0.750 \pm 0.028$ Macro-F1 for severity estimation and $0.925 \pm 0.013$ Macro-F1 for action recognition.  This improvement is likely driven by the rich spatial and contextual representations learned from large-scale pretraining, which are further refined through fine-tuning to better capture subtle movement impairments and variations in motion execution quality. Interestingly, when fine-tuning is not applied, the performance gap between video-based and IMU-based methods becomes considerably smaller, suggesting that wearable sensing remains a strong and viable low-cost alternative capable of capturing subtle movement alterations relevant to motor assessment. The multimodal late-fusion approach achieved the best overall performance, reaching $0.784  \pm 0.049$ Macro-F1 for severity estimation and $0.934 \pm 0.013$ Macro-F1 for action recognition. These results highlight the complementary nature of wearable inertial sensing and egocentric video, where IMUs provide robust low-level motion dynamics while visual information contributes contextual, spatial, and interaction-aware cues, resulting in a more comprehensive representation of complex movement patterns and activity execution characteristics.
Figure~\ref{fig:cm} presents the confusion matrices for severity estimation and action recognition obtained using IMU-based and video-based models. Residual confusions were primarily observed between activities exhibiting similar motion dynamics and between adjacent severity levels, highlighting the intrinsic difficulty of discriminating subtle variations in movement execution and impairment intensity.
\subsection{Single- and Multi-Sensor IMU Analysis}

Figure~\ref{fig:model_summary}~(Left) presents a comparative analysis of different IMU sensor configurations. The results reveal that wrist- and chest-mounted IMUs capture substantially more discriminative information for both severity estimation and action recognition than leg-based sensing alone. In particular, wrist-centered configurations achieved the strongest action recognition performance, while combining sensors provided the best overall trade-off across both tasks. Interestingly, incorporating the leg sensor yielded only marginal improvements despite increasing the feature dimensionality from 12 to 18 channels. One possible explanation is that participants in the proposed dataset often struggled to consistently mimic lower-limb and upper-body impairments simultaneously. Consequently, leg-mounted sensors may have captured less distinctive motion patterns compared to wrist and chest sensors. Nevertheless, these findings may not fully generalize to real clinical populations, where patients can exhibit persistent lower-limb impairment during resting, standing, or seated conditions, potentially increasing the contribution of leg-mounted sensing.

\section{Study Limitations}
A primary limitation of EgoInertia-MI is its reliance on simulated impairments from healthy volunteers. While this approach ensures a controlled, balanced, and reproducible dataset, simulated movements cannot fully replicate clinical realities. Pathological impairments are shaped by complex, interacting factors like rigidity, fatigue, compensatory behaviors, and medication effects; furthermore, our instruction-based severity levels do not directly map to clinically validated rating scales. Therefore, while this benchmark provides a structured framework for studying movement variations, evaluation on authentic clinical cohorts and longitudinal data remains essential to assess model transferability, robustness, and clinical relevance.

\section{Conclusion}
In this work, we introduced EgoInertia-MI, a multimodal benchmark dataset combining synchronized egocentric vision and wearable IMU sensing for motor impairment analysis. The benchmark spans 19 upper- and lower-body activities performed across multiple severity levels, offering standardized tasks for action recognition and impairment estimation. Our findings suggest that first-person visual observations encode rich cues for characterizing subtle variations in human motor behavior, while multimodal learning consistently achieves the best performance by exploiting the complementary nature of contextual vision and quantitative motion dynamics. Ultimately, EgoInertia-MI provides a scalable foundation for assessing the feasibility of integrating egocentric perception and wearable sensing within healthcare-oriented artificial intelligence. By bridging these domains, this work aims to foster future research toward naturalistic, privacy-aware, and accessible movement assessment systems. Future work will focus on extending the benchmark to authentic clinical populations and investigating self-supervised and foundation-model-based approaches for multimodal movement understanding in naturalistic environments.


\bibliographystyle{splncs04}
\bibliography{ref}

\end{document}